\documentclass[10pt,twocolumn,letterpaper]{article}

\usepackage{cvpr}
\usepackage{times}
\usepackage{epsfig}
\usepackage{graphicx}
\usepackage{amsmath}
\usepackage{amssymb}
\usepackage{color}
\usepackage{subfigure}

\usepackage[ruled]{algorithm2e}

\usepackage[breaklinks=true,bookmarks=false]{hyperref}

\cvprfinalcopy 

\def\cvprPaperID{11} 

\newcommand*{\affmark}[1][*]{\textsuperscript{#1}}

\ifcvprfinal\fi
\begin{document}

\title{Knowledge Representing: Efficient, Sparse Representation of Prior Knowledge for Knowledge Distillation}

\author{
Junjie Liu\affmark[1],
Dongchao Wen\affmark[1],
Hongxing Gao\affmark[1],
Wei Tao\affmark[1],\\
\affmark[1]{Canon Information Technology (Beijing) Co., LTD}\\
{\tt\small \{liujunjie, wendongchao, gaohongxing, taowei\}@canon-ib.com.cn}
\and
Tse-Wei Chen\affmark[2],
Kinya Osa\affmark[2],
Masami Kato\affmark[2]\\
\affmark[2]{Device Technology Development Headquarters, Canon Inc.}\\
{\tt\small twchen@ieee.org}
}

\maketitle
\thispagestyle{empty}

\begin{abstract}
  Despite the recent works on knowledge distillation (KD) have achieved
  a further improvement through elaborately modeling
  the decision boundary as the posterior knowledge,
  their performance is still dependent on the hypothesis that
  the target network has a powerful capacity (representation ability).
  In this paper,
  we propose a knowledge representing (KR) framework mainly
  focusing on modeling the parameters distribution as prior knowledge.
  Firstly,
  we suggest a knowledge aggregation scheme
  in order to answer how to represent the prior knowledge from teacher network.
  Through aggregating the parameters distribution from
  teacher network into more abstract level,
  the scheme is able to alleviate the phenomenon of
  residual accumulation in the deeper layers.
  Secondly,
  as the critical issue of
  what the most important prior knowledge is for better distilling,
  we design a sparse recoding penalty for constraining
  the student network to learn with the penalized gradients.
  With the proposed penalty,
  the student network can effectively avoid
  the over-regularization during knowledge distilling and converge faster.
  The quantitative experiments exhibit that
  the proposed framework achieves the state-of-the-arts performance,
  even though the target network does not have the expected capacity.
  Moreover,
  the framework is flexible enough for combining with
  other KD methods based on the posterior knowledge.
\end{abstract}






\section{Introduction}
The deep neural network has achieved the significant improvement
in different fields with years,
but it also requires higher computational and memory costs.
For the purpose to apply these networks to the real-time industrial tasks,
the neural network compression \cite{cheng2017survey} is arguably the most crucial strategy.
As for the network compression problem,
the typical solutions are designed
to \emph{slim} \cite{sandler2018mobilenetv2, zhang2018shufflenet} the network directly,
or quantify their
parameters distributions \cite{hubara2017quantized, jacob2018quantization, rastegari2016xnor},
and filter the
redundant layer dimensions \cite{he2017channel, thinet2017}.

\begin{figure}[t]
  \centering
  \includegraphics[width=.47\textwidth]{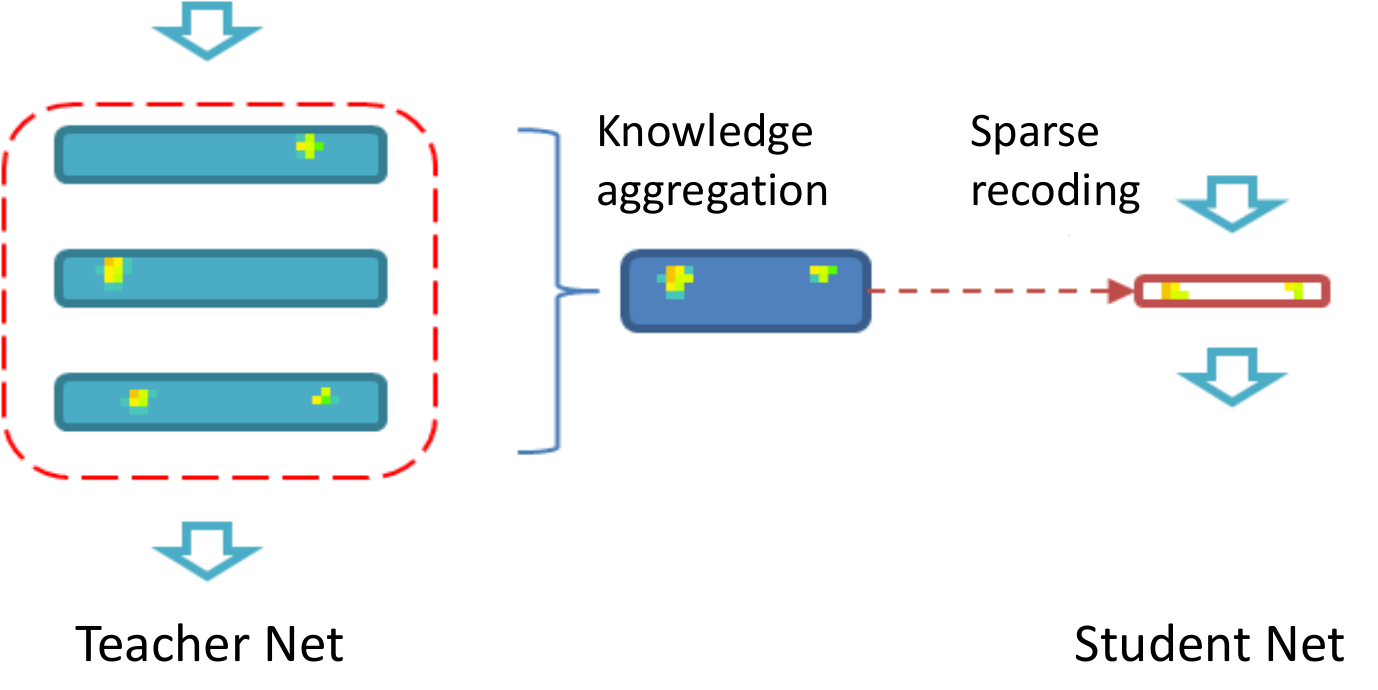}
  \caption{The pipeline of knowledge representing algorithm: The prior knowledge in teacher network is represented by the knowledge aggregation scheme into higher abstract level. Then the sparse recoding penalty is further used to regularize the gradients in student network for efficient learning these prior knowledge.}
  \label{fig:KD}
\end{figure}

In contrast to these techniques
which aim at directly compressing the network
while preserving its performance as much as possible,
an alternative solution is to preset a smaller target network as the student,
and employ the knowledge from the larger network as teacher
to improve student's performance.
Therefore,
knowledge distillation \cite{Hinton2015Distilling} (KD) is proposed.
The KD mainly assumes the samples distribution is anisotropy \cite{balan2015bayesian},
but annotations of the samples are not able to represent this intrinsic.
Based on the hypothesis,
these methods evaluate the samples in the teacher network
to produce the decision boundary as a strong posterior distribution,
and then use to regularize the gradients optimization of student network.
While this helps prevent the student network from being over-fitting,
the extra risk of non-convergence is introduced.

A possible solution is to refine
the posterior distribution from the teacher network,
in order to provide more valuable knowledge for better distilling.
The Neuron Selectively Transfer (NST) \cite{huang2017like} is proposed to
align the distribution selectively with the Maximum Mean Discrepancy (MMD) metric,
and the generative adversarial network with KD (KDGAN) \cite{wang2018kdgan}
is further used to produce a more robust decision boundary for student classifier.
However,
considering the student network which contains a very limited capacity - the representation ability,
this limitation gradually becomes a major bottleneck in network training
to further improve the performance of knowledge distillation.
In a word,
the fine-grained posterior distribution is usually underemployed.

With the constraint from network capacity,
an instinctive approach is to
introduce the parameters distribution \cite{denil2013predicting} from the teacher network
as the prior knowledge \cite{Romero2014FitNets, yim2017gift}.
For the typical one,
Romero et al. \cite{Romero2014FitNets} constructs the Hint layer
to estimate a parameters distribution with less filter numbers,
through using the intermediate features representation of the teacher,
and it uses these knowledge to guide the update of student parameters.
However,
the Hint layer suffers from the over-regularization if the teacher network is too deep.

In this paper,
we produce a KD solution mainly focusing on modeling the prior knowledge,
while avoiding the negative impacts from over-regularization,
and the solution is flexible enough,
for combining with other KD methods based on the posterior knowledge.
Specially,
we propose a knowledge representing (KR) framework,
which aims at representing the prior knowledge at more abstract level,
and taking full advantage of these knowledge.
For answering the question of
how to represent the prior knowledge,
a knowledge aggregation scheme is firstly suggested.
Inspired by the theory of optimal transportation \cite{Martin2017, Na2017A},
the scheme is designed to alleviate the phenomenon of
residual accumulation in the deeper layers.
Then,
as for the most critical issue of
what the dominant prior knowledge is for better distilling,
a sparse recoding penalty is proposed.
Through employing a learnable threshold in the penalty,
it can enhance the gradients of dominant neurons
and smooth inactive ones.
With these two proposed terms,
the proposed framework can prompt the student network
to preserve the key features of teacher network,
even without a strong representation ability.

Our paper makes the following contributions:
\begin{itemize}
\item A new penalty is proposed to constrain the optimization of knowledge distillation. It helps the student network to avoid the over-regularization and converge faster. Moreover, the penalty can be further applied on other network optimization problems.
\item A new scheme is suggested for aggregating the prior knowledge. It is able to produce more abstract features and alleviate the phenomenon of residual accumulation.
\item According to the proposed framework, the more flexible architecture is allowable for both teacher and student network, without the constraints from model depths or filter scales.
\end{itemize}
\section{Related Work}

The latest deep networks are usually accompanied with
carefully designed modules \cite{denseNet2017, he2016deep}
and enormous parameters.
Though the performance of targeted tasks is obviously being improved,
the computation and memory cost gradually become the challenge
to employ these networks
in real-life applications \cite{jacob2018quantization, Shen2018Towards}.
Comparing to the traditional neural network compression methods \cite{cheng2017survey}
which focus on compressing the original network directly,
a solution with the knowledge distillation
to compress the deep network attracts more attention from research community in recent years,
such as in the tasks of image recognition \cite{yim2017gift},
object detection \cite{GuobinChen_2017_NIPS, luo2016face},
or recommender systems \cite{Guorui_2018_AAAI},
as the flexibility to obtain an arbitrary architecture of target network.
In summary,
the KD methods can be categorized into two main groups:

1) \emph{Distilling the posterior distribution from training data:}
Considering the possibility to extract the knowledge
in an ensemble (teacher) into a single model (student),
Hinton et al. \cite{Hinton2015Distilling} introduces
the idea of knowledge distillation as a regularizer.
Through employing a penalized version \cite{BSS2019, huang2018data, zheng2017CKD}
of final features of the teacher network,
a joint learning is processed with the knowledge from posterior distribution.
For refining the posterior distribution to
provide more valuable knowledge,
the Neuron Selectively Transfer (NST) \cite{huang2017like} is proposed to
align the distribution selectively with the Maximum Mean Discrepancy (MMD) metric.
Furthermore,
considering the sample bias is unavoidable,
the generative adversarial networks for knowledge distillation (KDGAN) \cite{wang2018kdgan}
is further used to produce a more robust posterior distribution for student classifier.
However,
these methods haven't take the capacity of student network into consideration,
so the fine-grained posterior distribution is underemployed.


2) \emph{Distilling the prior distribution from model parameters:}
An alternative approach is to
introduce the parameters distribution from teacher network
as the prior knowledge \cite{Romero2014FitNets, yim2017gift, Zagoruyko2017Paying}.
Romero et al. \cite{Romero2014FitNets} designs the Hint layer to
estimate the parameters distribution
by using the intermediate hidden layers from the teacher,
and used the Hint layer to guide the distillation.
Net2Net \cite{Chen2015Net2Net} suggests a function-preserving transform
for extracting the prior knowledge from teacher network
to initialize the parameters of the student network.
And Yim et al. \cite{yim2017gift} suggests
a representation operator named FSP matrix.
It uses not only the parameters distribution
but the intermediate features from the neighbor layers.
However,
these methods either are constrained by the depth of teacher network,
or suffer from the over-regularization.

\section{Method}
For obtaining a student network that
faithfully preserves the key representation ability of the teacher,
Sec. \ref{3dot1} presents the objective function
of the knowledge representing framework.
Accordingly,
we firstly answer the key problem of
what the most important prior knowledge is for distilling in Sec. \ref{3dot2},
through introducing the mathematical expression of the sparse recoding penalty.
Then,
we suggest how to represent the prior knowledge from the teacher network,
with a knowledge aggregation scheme in Sec. \ref{3dot3}.
Finally, Sec. \ref{3dot4} shows the optimization procedure
of the objective function.

\begin{algorithm}[b]
    \caption{Training the Knowledge Representing Algorithm}
    \KwIn{Weights $W^t$ of teacher network and $W^s$ of student network}
    \KwOut{Aggregation knowledge $\tilde{W^t}$ and optimized weights of student network $W^s$}
    \textbf{Initialization:} $\tilde{W^t}$, $W^s$, $MaxIter$\;
    \While{less than the MaxIter}{
        \textbf{Optimizing} $\tilde{W^t}$ \textbf{with} $W^t$\;
            Knowledge aggregation for teacher in Eq. \ref{eq:interpret-KD-further-revised-solved} \;
        \textbf{Optimizing} $W^s$ \textbf{with} $\tilde{W^t}$\;
            Sparse recoding for student in Eq. \ref{eq:interpret-KD-revised-split2} \;
    }
\label{alg:frame}
\end{algorithm}

\subsection{Knowledge Representing} \label{3dot1}
As one of the most typical feature representation technique,
the deep model produces the decision boundary through
modeling the data distribution with the parameters in layers.
Given a trained decision boundary $y^{t}(x, W^t)$,
where $y^t$ is generated by teacher network with
data distribution $x$ and the parameters $W^t$,
the objective of knowledge distillation is
to find the parameters $W^s$ for the student network.
Specially,
with the $W^s$ and $x$,
the $y^s$ from student network
is jointly optimized with the $y^t$.
Through minimizing the dissimilarity of two decision boundaries,
the objective function of knowledge distillation is defined as:

\begin{equation}
\underset{W^s}{\arg\min} \sum_{i=1}^{N}\pounds(y_{i}^{t}(x_i, W^t), y_{i}^{s}(x_i, W^s)) + \lambda \Phi(W^s)
\label{eq:interpret-KD}
\end{equation}

where $\pounds$ represents the metric for evaluating
the similarity between the $y^t$ and $y^s$,
and the cross entropy, KDGAN \cite{wang2018kdgan}, or NST \cite{huang2017like} are allowable.
Different from the KD methods only evaluating the decision boundary,
we further introduce a penalty $\Phi(\cdot)$ in Eq. \ref{eq:interpret-KD},
in order to measure the representation ability of student network.
However,
if the representation ability of student network is weak,
the fine-grained posterior distribution will be underemployed.
Then,
we extend the objective function Eq. \ref{eq:interpret-KD}
through further introducing the prior knowledge $W^t$ from the teacher network,
and the objective function is:

\begin{equation}
\begin{split}
\underset{W^s, \tilde{W^t}}{\arg\min} \sum_{i=1}^{N}&\pounds(y_{i}^{t}(x_i, W^t), y_{i}^{s}(x_i, W^s)) + \pounds(\tilde{W^t}, W^s) \\
&+ \gamma \Psi(\tilde{W^t}, W^t) + \lambda \Phi(W^s)
\label{eq:interpret-KD-revised}
\end{split}
\end{equation}

Instead of directly employing the parameter distributions $W^t$
from the teacher network as prior knowledge,
we firstly represent these distributions as more abstract level,
and a knowledge aggregation scheme $\Psi(\cdot)$ is suggested
to aggregate $W^t$ into the $\tilde{W^t}$.
With the prior knowledge $\tilde{W^t}$,
the $\pounds(\tilde{W^t}, W^s)$ is used to guide the
update of parameters distributions $W^s$ for the student.
Moreover,
we propose a sparse recoding penalty to specify the $\Phi(\cdot)$.
Through enhancing the magnitude of dominant gradients and filtering the inactive ones,
the optimizer no longer requires the parameters distribution $W^s$ of student network
to strictly close to the teacher one,
and prompts the student network to firstly learn with the most valuable knowledge.
In summary,
the optimization procedure is represented in Algorithm \ref{alg:frame},
and we leave over the details in following sections.

\begin{figure}[t]
  \centering
  \subfigure[]{\includegraphics[width=.23\textwidth]{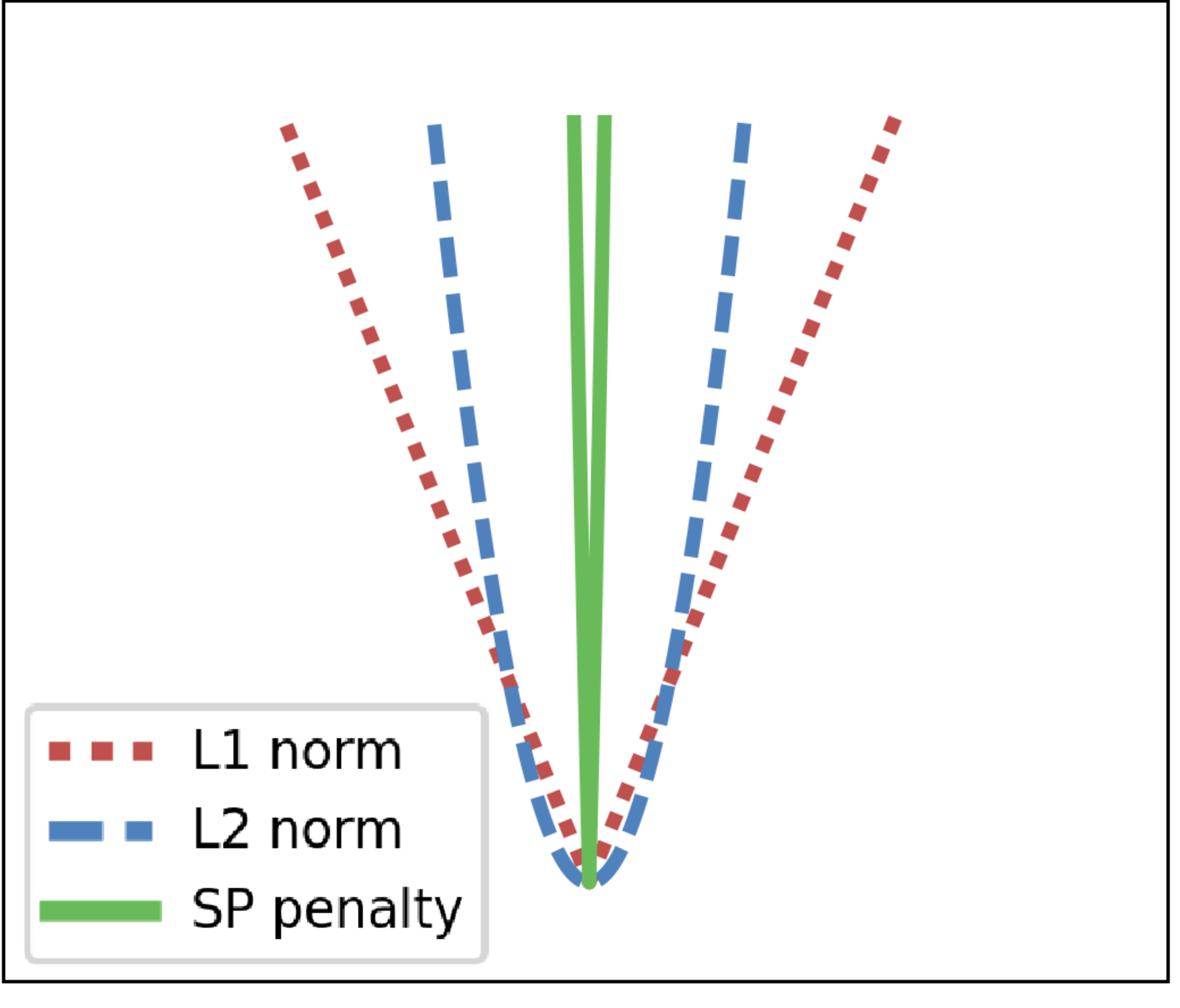}}
  \subfigure[]{\includegraphics[width=.23\textwidth]{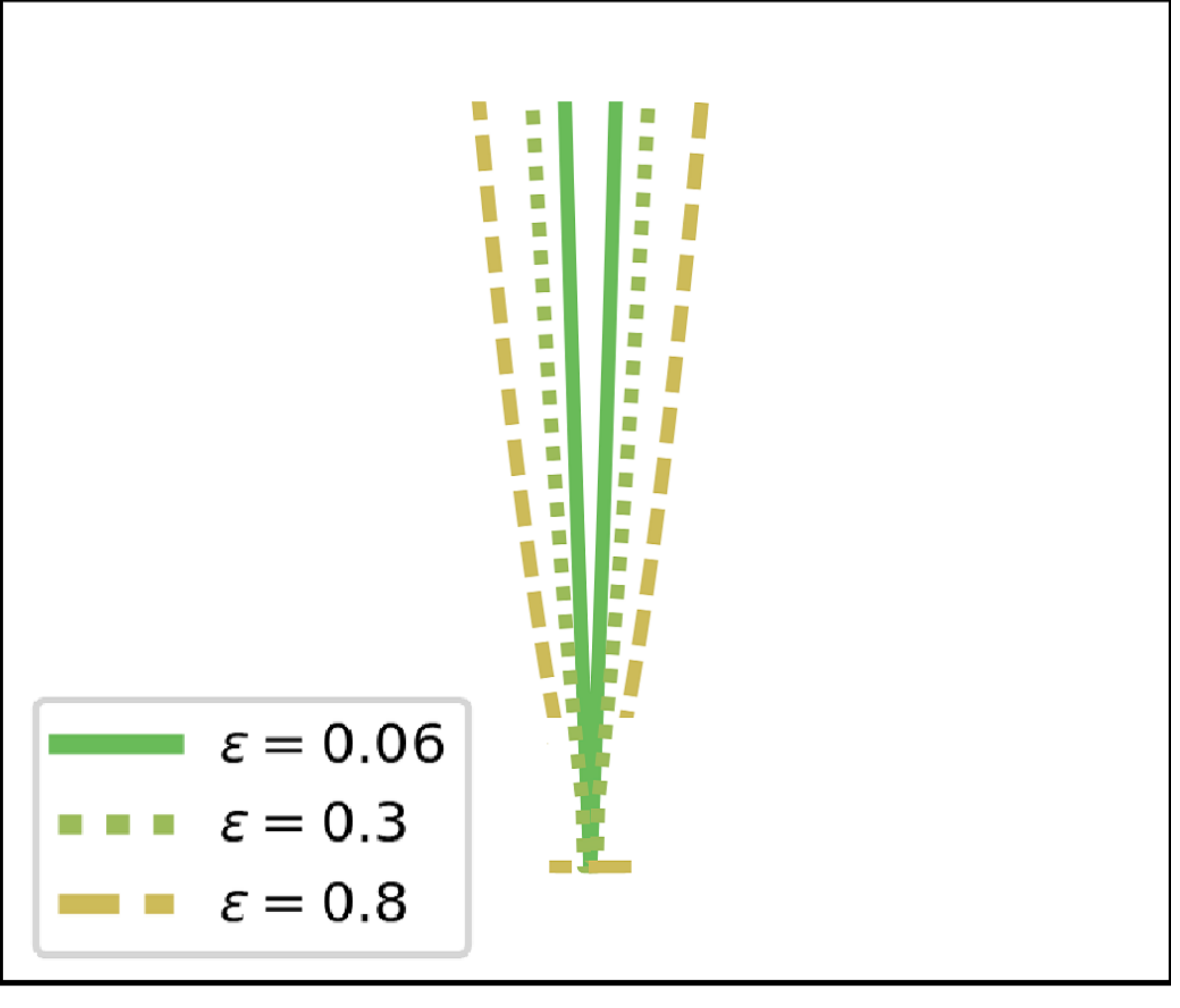}}
  \caption{The properties of proposed sparse recoding penalty (best viewed in color):
  (a) the sparse recoding penalty is able to approximate more strict sparseness;
  (b) the properties with different setting of $\varepsilon$.}
  \label{fig:sp-loss}
\end{figure}

\subsection{Sparse Recoding Penalty} \label{3dot2} 
As demonstrated by previous works \cite{wen2016learning, wu2017beyond, zhang2016l1},
prompting the neurons connection being sparse is beneficial for
obtaining a well generalization ability.
However,
such penalties are designed to directly clip the parameters distribution,
and the extra risk of over-regularization is introduced.
After we analyze the distribution of prorogated gradients in the previous KD methods,
we found that major reason for the convergence of oscillatory
is that the gradients are not discriminative enough,
especially in the student network with a weak representation ability.

Therefore,
we propose a sparse recoding penalty $\Phi(\cdot)$,
which can penalize the prorogated gradients during the training of deep network.
Given an input parameters tensor $W$,
it enhances the high gradients $g_j$ of dominant neurons,
and filters the low gradients of inactive neurons.
The function is defined as:

\begin{equation}
\begin{split}
\Phi(W) = \sum_{j}\Phi_0(g_{j})
\label{eq:sp-loss}
\end{split}
\end{equation}

where

\begin{equation}
\Phi_0(g)=
\begin{cases}
\frac{1}{\varepsilon}(|g| + g^2), &if|g| \geq \varepsilon \\
0, &otherwise
\end{cases}
\label{eq:sp-loss-cases}
\end{equation}

where $\Phi_0(\cdot)$ is a piecewise function that
enhances the gradients when $|g| \geq \varepsilon$,
and smooths the $|g|$ by zero in others.
The $\varepsilon$ is a learnable threshold within the update of gradient optimization,
and it is initialized with the mean value of parameters distribution.
For fairly comparing with other penalties,
the Fig.\ref{fig:sp-loss} shows
the curves of $\Phi(\cdot)$ by comparing with the $L_1$ and $L_2$ norms.
It exhibits that $\Phi(\cdot)$ is a more strict sparse constraint.
Moreover, with different parameter setting,
properties of the sparse recoding penalty are shown in the figure,
and we leave over the further discussion in experiments.

\subsection{Deep Knowledge Aggregation} \label{3dot3}
For representing the prior knowledge as more abstract level,
we design a deep knowledge aggregation scheme
through stacking the neighbor layers in a very deep network.
Specially,
with the analysis of prior knowledge distilling in previous methods,
we notice that the optimization errors between two networks will be accumulated from layers,
since the higher layer in teacher network usually contains a strong representation ability.
However,
the situation is simply regarded as the phenomenon of gradient vanishing,
and cause an over-regularization if the teacher network is too deep.
So we name this phenomenon as the residual accumulation,
and the proposed scheme will mainly considers this phenomenon.
Based on the theory of optimal transportation \cite{Martin2017, Na2017A},
the scheme try to reduce the residual accumulation during gradient optimization,
through minimizing the inter-domain transportation cost.
Given a $P_1$ and $P_2$ being two distribution space
with probability measures $\mu$ and $\nu$ respectively,
the transportation $T$ preserving $P_1 \rightarrow P_2$ has equal total measure

\begin{equation}
\mu(T(p_1)) = \nu(p_2)
\label{eq:layer-ag}
\end{equation}

where $p_1$ and $p_2$ is any measurable subset of $P_1$ and $P_2$.
Then the total transportation cost for sending $p_1 \subset P_1$ to $p_2 \subset P_2$
by transportation cost $\tau(p_1,p_2)$
can be defined by

\begin{equation}
\min_{T:P_1 \rightarrow P_2}\int_{P_1}\tau(p_1,T(p_1))d\mu(p_1)
\label{eq:layer-ag-tp}
\end{equation}

With minimizing the total transportation cost,
the distribution $P_2$ progressively approximates $P_1$ on measures $\mu$.
Assuming a series of neighbor layers ${\ell_k, ..., \ell_n}$ as set $\textit{\L}_k^n$,
for sending parameter distribution $W_{\textit{\L}_k^n}$ to $\tilde{W}$
with measurable subset $w \subset W_{\ell_k, ..., \ell_n}$,
the deep knowledge aggregation scheme merges the neighbouring layers
to form the higher abstract parameters knowledge.
In this case,
the function $\Psi(\cdot)$ is formulated as

\begin{equation}
\Psi(\tilde{W}, W_{\textit{\L}_k^n}) = \min_{T:W_{\textit{\L}_k^n} \rightarrow \tilde{W}}\int_{W_{\textit{\L}_k^n}}\tau(w,T(w))d\mu(w)
\label{eq:layer-ag-details}
\end{equation}


\subsection{Optimization} \label{3dot4}
Instead of directly optimizing the proposed objective function,
we design an joint optimization method as the alternative solution.
In details, our method uses two stages optimization
to alternatingly solve the Eq. \ref{eq:interpret-KD-revised}.

\paragraph{Optimizing $\tilde{W^t}$ with $W^t_{\textit{\L}_k^n}$}
Given an elaborate teacher network with parameter distribution $W^t_{\textit{\L}_k^n}$,
we first aggregate the knowledge $\tilde{W^t}$ with $T({W^t_{\textit{\L}_k^n}})$ in here
as:

\begin{equation}
\begin{split}
&\underset{\tilde{W^t}}{\arg\min} \sum_{i=1}^{N}\pounds(y_{i}^{t}(x_i, W^t), y_{i}^{s}(x_i, W^s))\\
&+ \gamma \int_{W_{\textit{\L}_k^n}^t}\tau(w^t,T(w^t))d\mu(w^t)
\label{eq:interpret-KD-revised-solve}
\end{split}
\end{equation}

As the Eq. \ref{eq:interpret-KD-revised-solve} involves a transportation cost
and the definition of probability measures,
it is difficult to directly integrate with gradient descent optimizer.
In this case,
we use the feature representation $\textit{F}_W$ as an approximation probability measures,
which means the set of features maps $\textit{F}$ generated by parameters set $W$.
If the transportation cost $\tau(\cdot)$ is defined as the simple $L_2$ distance,
we revise the Eq. \ref{eq:interpret-KD-revised-solve} as:

\begin{equation}
\begin{split}
\underset{\tilde{W^t}}{\arg\min} \sum_{i=1}^{N}&\pounds(y_{i}^{t}(x_i, W^t), y_{i}^{s}(x_i, W^s))\\
&+ \gamma \mu(W^t_{\textit{\L}_k^n})\|\textit{F}_{W^t_{\textit{\L}_k^n}} - \textit{F}_{\tilde{W^t}}\|_2
\label{eq:interpret-KD-further-revised-solved}
\end{split}
\end{equation}

where $\gamma$ is a predefined parameter to control the penalty from optimal transportation.
The $\mu(W^t_{\textit{\L}_k^n})$ as a measures function is used to
penalize more on the layer with higher accumulation error,
and the standard deviation is employed here.
Moreover, we remove the part of terms during the derivation
for Eq. \ref{eq:interpret-KD-revised-solve} for fast computation.
Then, the solution of $\tilde{W^t}$ can be obtained by gradient descent optimization.

\paragraph{Optimizing $W^s$ with $\tilde{W^t}$}
Given an aggregate knowledge $\tilde{W^t}$,
our goal here is further to solve the $W^s$ on student network with sparse recoding penalty,
as:
\begin{equation}
\begin{split}
\underset{W^s}{\arg\min} \pounds(\tilde{W^t}, W^s) + \lambda \sum_{j}\Phi_0(g_{j}^{s})
\label{eq:interpret-KD-revised-split2}
\end{split}
\end{equation}

where $\Phi_0(g_{j}^{s})$ is designed for
prompting the student network to firstly learn with the penalized gradients,
and the parameter $\lambda$ is predefined to control
the importance of the sparse recoding penalty.

Instead of directly solving the global optimum
for objective function Eq. \ref{eq:interpret-KD-revised},
the two sub-objective functions
Eq. \ref{eq:interpret-KD-further-revised-solved}
and Eq. \ref{eq:interpret-KD-revised-split2}
are designed to
overcome the conflict between
optimizing the prior knowledge and posterior knowledge simultaneously.
Through alternatively minimizing the
distribution dissimlarity $\pounds(\tilde{W^t}, W^s)$
and $\pounds(y^{t}, y^{s})$,
the optimization for Eq. \ref{eq:interpret-KD-revised} is regarded as
an joint optimization procedure.
Once the posterior knowledge is dominant during optimization,
the optimizer for prior knowledge will penalize the total loss more,
and the opposite is also.
The gradient is only allowed to descend
on the direction that makes both two optimizers are optimal.

\section{Experiments}
\begin{figure}[htb]
  \centering
  \includegraphics[width=.47\textwidth]{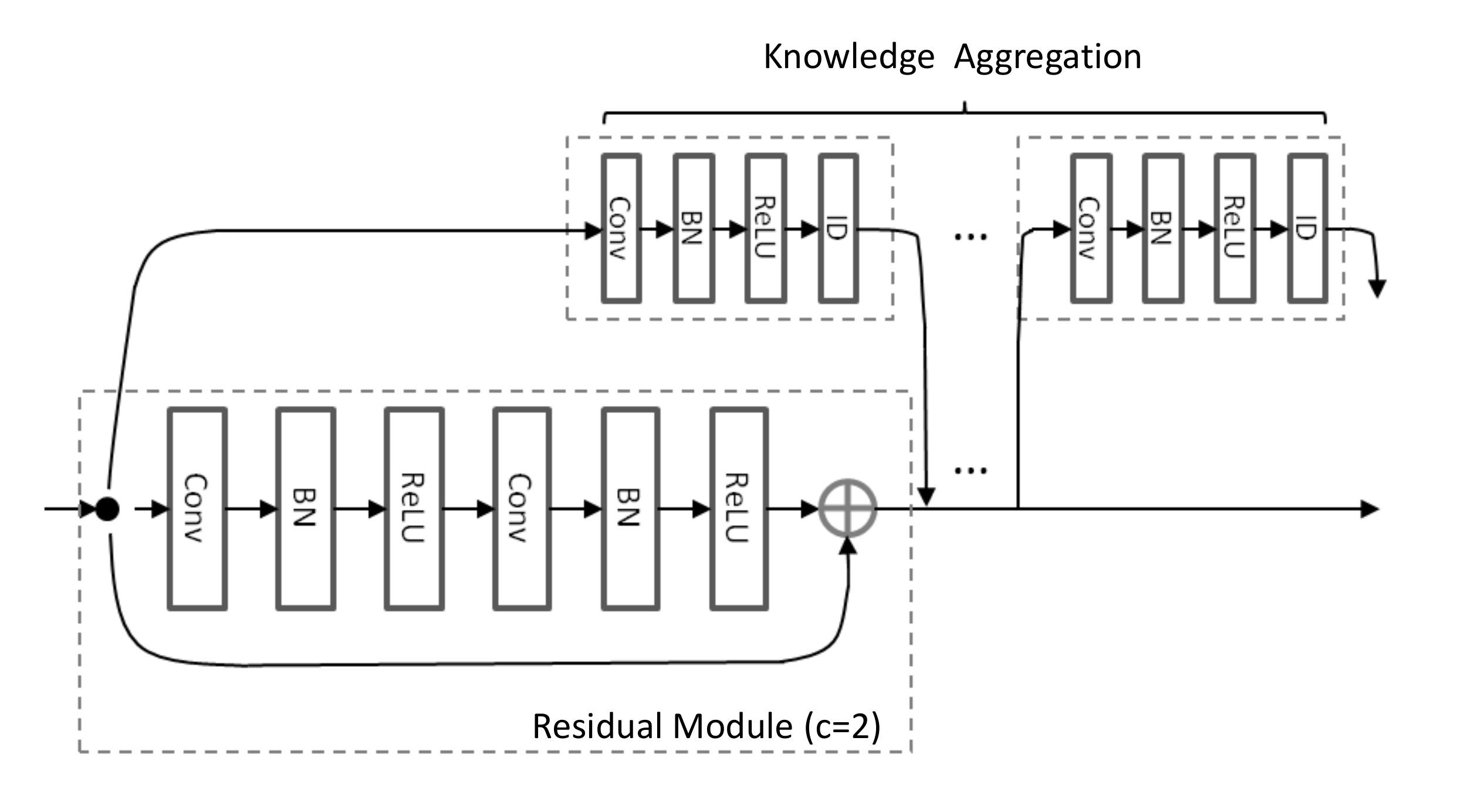}
  \caption{The excerpt of proposed framework on the residual network.}
  \label{fig:residual_module}
\end{figure}

In this section,
we evaluate the proposed knowledge distillation framework with several benchmark datasets.
For the base of experiments,
we use the deep residual network \cite{he2016deep}
as the network architecture,
and the excerpt of the proposed framework in this architecture
is shown in Fig. \ref{fig:residual_module}.
The $c$ in residual module means the number of aggregated convolution layers.
For the problem of optimizing these layers with different spatial scales,
the identity mapping (ID) layer \cite{Yu2018Learning} is employed also.
To ensure a fair comparison,
the same data augment strategies are used.
Moreover,
we employ the similar settings of learning rates,
optimization iterations and
computation precision (32 float points).
The implementation details
will be shown in corresponding subsections.

In Sec. \ref{sec_penalty_analysis},
through comparing with the typical penalties,
the property of the sparse recoding penalty is analysed.
Then,
through comparing with the state-of-the-arts,
we evaluate the performance of student networks in general image recognition tasks,
and further explore their generalization ability in a revised dataset TCIFAR-100,
as described in Sec. \ref{sec_performance_analysis}.
Finally,
the discussions about the optimization procedure of the proposed framework is shown in Sec. \ref{sec_optimization_analysis}.

\subsection{Analysis of Proposed Penalty}\label{sec_penalty_analysis}
As for the sparse recoding penalty,
its property through comparing with typical methods is analysed,
and we further explore the reason of why the proposed penalty is able to boost
the convergence of knowledge distilling.
Based on the experiment result,
we address that
the proposed penalty can be applied on
other network optimization problems
if the gradients distribution is not discriminative enough.

\begin{figure}[ht]
  \centering
  \includegraphics[width=.42\textwidth]{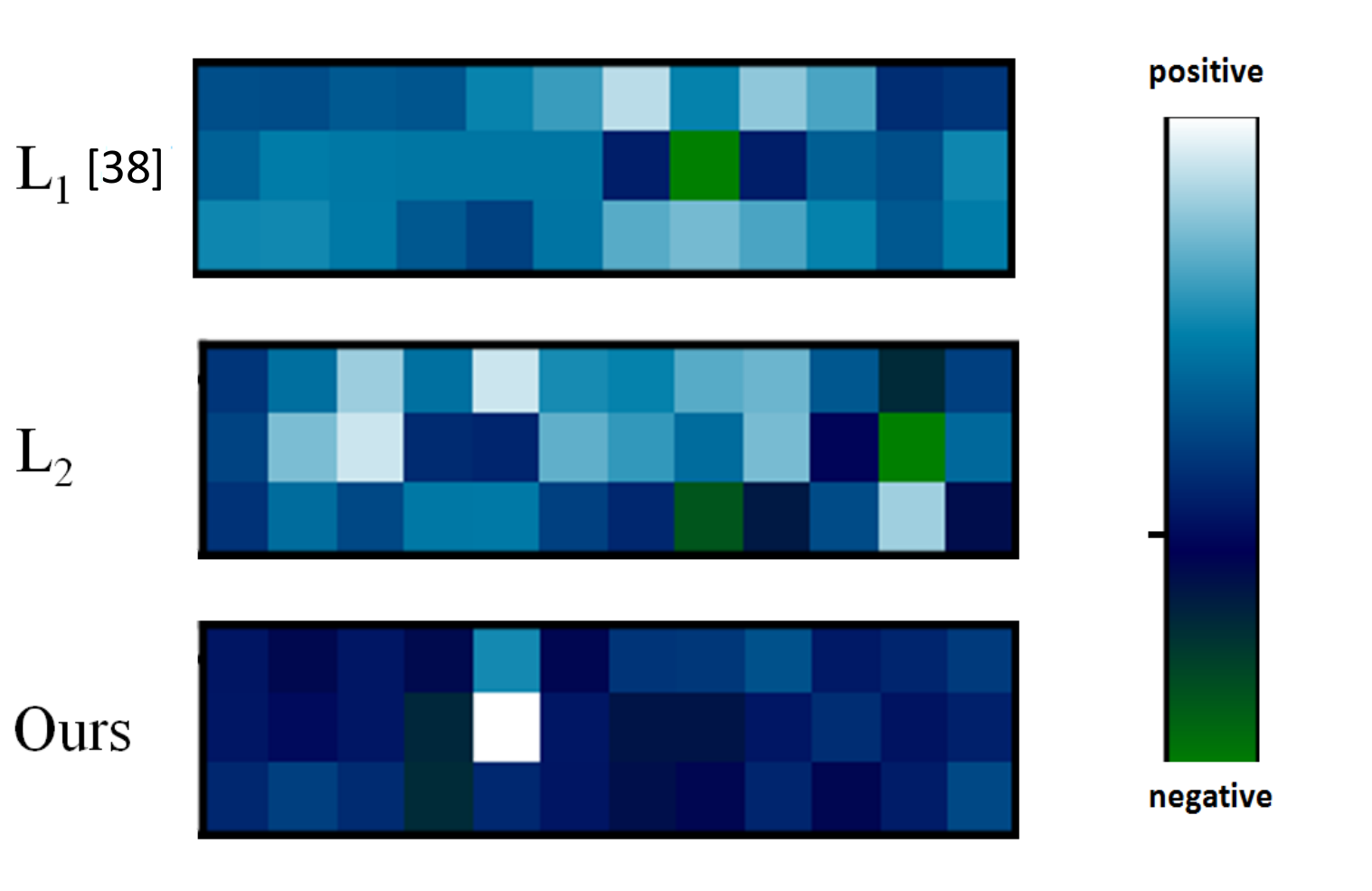}
  \caption{With the penalized gradients, the sparse recoding penalty is also to produce
  more discriminative parameters distribution (best see in color).}
  \label{fig:kernel-analysis}
\end{figure}

\paragraph{Penalty Property}
Given a specific parameters distribution,
the traditional penalties \cite{wen2016learning, zhang2016l1} form a convex function
and obtain the maximal reward in the unique extreme.
It penalizes the parameter with higher value to reduce the total loss,
for encouraging the value of parameter to close to 0.
In contrast to these methods,
the sparse recoding penalty is designed to penalize the gradients directly.
For the propagated gradients,
it filters the gradients with an equal reward within the learnable threshold,
in order to slow down the update of inactive neurons.
For the gradients out of the threshold,
it boosts the update to highlight the dominant neurons.
For validating our hypothesis,
we visualize the convolutional kernels with the constraint
by different penalties in image recognition tasks.
The Fig. \ref{fig:kernel-analysis} shows that the sparse recoding penalty
can prompt the parameters distribution of the network to be more sparse,
through directly regularizing the optimized gradients.

\begin{figure}[b]
  \centering
  \includegraphics[width=.47\textwidth]{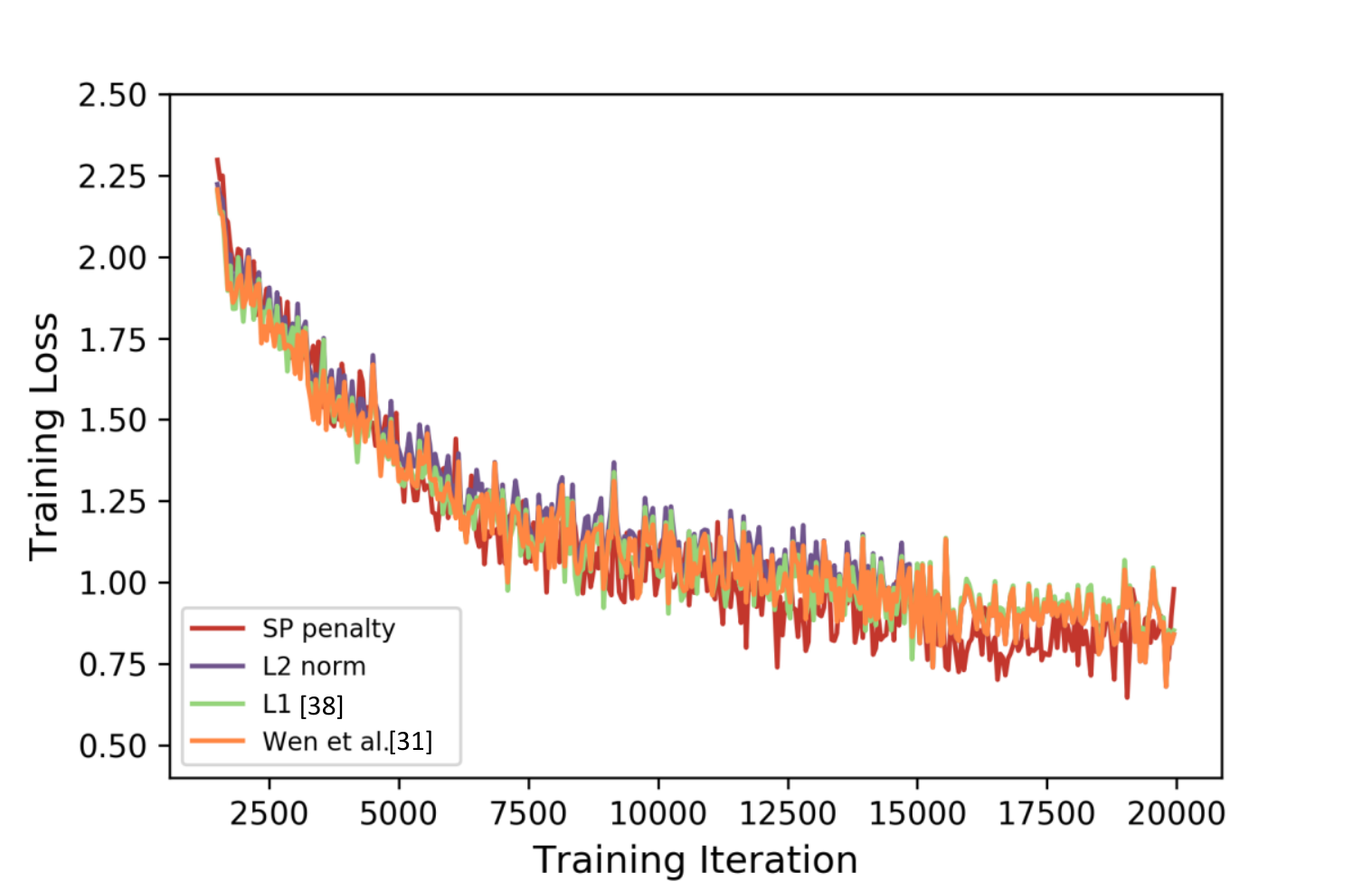}
  \caption{Convergence speed; the traditional penalties \cite{wen2016learning, zhang2016l1} and the sparse recoding penalty (best see in color).}                                                \label{fig:analysis_convergence}
\end{figure}

\paragraph{Convergence}
We have observed fast convergence in our experiment result.
In Fig. \ref{fig:analysis_convergence},
it illustrates the training loss on MNIST over the beginning 20,000 iteraitons.
The student network with sparse recoding penalty
is better than the traditional penalties.
We think one possible reason is that
the proposed penalty is designed to penalize the gradients firstly,
so it can produce a bigger step for gradients descent
in the beginning of network training.
Moreover,
we evaluate the different types for initializing the parameters distribution in the experiment,
and we also found the similar conclusion.

\subsection{Performance Analysis}\label{sec_performance_analysis}
In this section,
we firstly conduct the experiments
in the image recognition task
on CIFAR-10, CIFAR-100 \cite{Krizhevsky2009} and ILSVRC 2012 \cite{Deng2009},
in order to evaluate the performance of
the proposed knowledge representing framework with the state-of-the-arts.
Then,
we design a TCIFAR-100 based on CIFAR-100,
for further verifying their generalization ability.
As the focus of this experiment is analysing
the performance of student network with a small capacity,
so we reserve the comparison on different tasks as future works.

\subsubsection{CIFAR-10}
The CIFAR-10 is an image recognition dataset \cite{Krizhevsky2009}
which includes 50,000 training images and 10,000 test images,
and per training class has 5,000 images while test class has 1000 images.
For all images, they store in RGB format with size of $32 \times 32$.
We use a trained teacher network with 26 layers,
which is structured as 5 residual modules.
For student network,
it contains 8 layers with 2 residual modules,
which has roughly $1\//3$ parameters of the teacher.
In details,
with the same parameters settings and training strategies,
we reduce about $1\//3$ number of the filters on each layer for the student network,
in order to evaluate the case if the target network contains a weak representation ability.
And we set the $c$ of knowledge aggregation as 3,
which aggregates each three layer of teacher network into higher abstract level
for one layer in student network.

\begin{table}[h]
\begin{center}
\begin{tabular}{l|c|c}
\hline
 & Accuracy & Params\\
\hline
Teacher ResNet-26 & 91.91 & $\sim$ 0.36M\\
Student ResNet-8 (Original) & 87.91 & $\sim$ 0.12M\\
FitNet \cite{Romero2014FitNets} & 88.57 & $\sim$ 0.12M\\
FSP \cite{yim2017gift} & 88.70 & $\sim$ 0.12M\\
Proposed-Dense & 89.11 & $\sim$ 0.09M\\
Proposed & \textbf{90.65} & $\sim$ 0.09M\\
NTS \cite{huang2017like} & 88.98 & $\sim$ 0.12M\\
KDGAN \cite{wang2018kdgan} & 88.62 & $\sim$ 0.12M\\
Proposed + KDGAN \cite{wang2018kdgan} & \textbf{91.35} & $\sim$ 0.09M\\
\hline
\end{tabular}
\end{center}
\caption{ResNet-8 in CIFAR-10 Classification rates(\%). Proposed: the KR framework. Proposed-Dense: the KR framework but removing the sparse recoding penalty.}
\label{tab:cifar10}
\end{table}

In Tab. \ref{tab:cifar10},
it summarizes the obtained results.
Based on the proposed framework,
the student network which contains less parameters
wins the methods \cite{Romero2014FitNets, yim2017gift}
focusing on prior knowledge with a significant improvement.
For the state-of-the-arts \cite{huang2017like, wang2018kdgan} by modeling the posterior knowledge,
the proposed framework also achieves the comparable performance.
For the self-comparison,
we remove the sparse recoding penalty in KR framework and name it as the KR-Dense.
And the experiment proves
the importance to sparsely penalize the gradients during the distilling optimization,
if the student network only has a small capacity.
Besides,
through combining with the KDGAN \cite{wang2018kdgan},
a further improvement confirms that
our method is flexible for the extension.

\subsubsection{CIFAR-100}
The CIFAR-100 is an augmented version of CIFAR-10.
It contains the same amount of images and size of CIFAR-10,
which includes 50,000 training images and 10,000 test images,
so only has 100 samples per class.
Similar the setting to CIFAR-10,
we use a trained teacher network with 32 layers as 6 residual modules,
and student is composed of 14 layers as 3 residual modules.
Besides,
the reduction of about $1\//3$ filter number is still used,
and $c$ is set as 3.

Tab. \ref{tab:cifar100} shows results of student network with evaluated methods.
Though the proposed method achieves the comparable performance
than the state-of-the-arts \cite{huang2017like, wang2018kdgan} with less parameters,
the improvement for our method is not obvious.
We think one possible reason is that
the ResNet-14 has a stronger representation ability that the ResNet-8.

\begin{table}[t]
\begin{center}
\begin{tabular}{l|c|c}
\hline
 & Accuracy & Params\\
\hline
Teacher ResNet-32 & 64.06 & $\sim$ 0.46M\\
Student ResNet-14 (Original) & 58.65 & $\sim$ 0.19M\\
FitNet \cite{Romero2014FitNets} & 61.28 & $\sim$ 0.19M\\
FSP \cite{yim2017gift} & 63.33 & $\sim$ 0.19M\\
Proposed Method & 63.95 & $\sim$ 0.17M\\
NTS \cite{huang2017like} & 63.78 & $\sim$ 0.19M\\
KDGAN \cite{wang2018kdgan} & \textbf{63.96} & $\sim$ 0.19M\\
Proposed Method + KDGAN \cite{wang2018kdgan}  & \textbf{63.98} & $\sim$ 0.17M\\
\hline
\end{tabular}
\end{center}
\caption{ResNet-14 in CIFAR-100 Classification rates(\%). With the similar network architecture, we further reduce the output channels in each layer for saving the total parameters.}
\label{tab:cifar100}
\end{table}

\subsubsection{ILSVRC 2012}
The ILSVRC 2012 classification challenge
involves the recognition task to classify one image into 1,000 leaf-node categories
in the ImageNet hierarchy \cite{Krizhevsky2012}.
It has about 1.2 million images for training,
50,000 for validation and 100,000 testing images.
Although training the very deep network on such enormous datasets
to achieve satisfied performance has been a solvable issue,
how to obtain the comparable performance with
a tiny network by the knowledge distillation
still confuses the research community,
especially for the methods \cite{luo2016face, Romero2014FitNets, yim2017gift} with prior knowledge.
We think the major reason is that
the depth of teacher network in ILSVRC 2012 is very deep,
so the student network in these methods seriously suffers from the over-regularization.

Tab. \ref{tab:imagenet1} shows the errors of Top-1 and Top-5.
With the $c$ which is set as 4 in knowledge aggregation scheme,
we found the situation of over-regularization is alleviated,
and it prompts the KR framework to achieve the better performance.

\begin{table}[h]
\begin{center}
\begin{tabular}{l|c|c}
\hline
 & Top-1 & Top-5 \\
\hline
Teacher ResNet-101 & 22.68 & 6.58 \\
Student Inception-BN \cite{Bn2015} & 25.74 & 8.07 \\
FitNet \cite{Romero2014FitNets} & 25.30 & 7.93 \\
NTS \cite{huang2017like} & 24.34 & 7.11 \\
KDGAN \cite{wang2018kdgan} & 24.11 & 6.98 \\
Proposed Method & \textbf{23.47} & \textbf{6.85} \\
Proposed Method + KDGAN \cite{wang2018kdgan} & \textbf{23.18} & \textbf{6.79} \\
\hline
\end{tabular}
\end{center}
\caption{ImageNet Classification errors (Top-1 and Top-5\%). }
\label{tab:imagenet1}
\end{table}

\subsubsection{Generalization Ability}
\begin{figure}[ht]
  \centering
  \includegraphics[width=.4\textwidth]{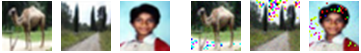}
  \caption{\emph{left} CIFAR-100; \emph{right} TCIFAR-100}
  \label{fig:tcifar-100ep}
\end{figure}

We further explore the generalization ability of
previous methods and the proposed framework.
Based on the data resource from CIFAR-100,
we reproduce the CIFAR-100 as the TCIFAR-100
with the data distortion strategies.
In details,
each image in training and test set is distorted by the artifacts,
from a gaussian distribution ($\sigma$ = 1) with the random sample.
The Fig. \ref{fig:tcifar-100ep} shows the examples.
In Tab. \ref{tab:tcifar100},
it shows the proposed framework
achieves a significant improvement than state-of-the-arts.
We believe the KR framework is able to produce
a student network with stronger generalization ability,
since the joint optimization prevents the optimizer
from being trapped in local extremum.

\begin{table}[ht]
\begin{center}
\begin{tabular}{l|c|c}
\hline
 & Accuracy & Params\\
\hline
Teacher ResNet-32 & 61.25 & $\sim$ 0.46M\\
Student ResNet-14 (Original) & 54.37 & $\sim$ 0.19M\\
FitNet \cite{Romero2014FitNets} & 56.77 & $\sim$ 0.19M\\
FSP \cite{yim2017gift} & 57.31 & $\sim$ 0.19M\\
Proposed Method & \textbf{60.03} & $\sim$ 0.17M\\
NTS \cite{huang2017like} & 57.88 & $\sim$ 0.19M\\
KDGAN \cite{wang2018kdgan} & 58.15 & $\sim$ 0.19M\\
Proposed Method + KDGAN \cite{wang2018kdgan}  & \textbf{60.33} & $\sim$ 0.17M\\
\hline
\end{tabular}
\end{center}
\caption{ResNet-14 in TCIFAR-100 Classification rates(\%). The transformed CIFAR-100 dataset is reproduced by the CIFAR-100.}
\label{tab:tcifar100}
\end{table}

\subsection{Optimization Discussion}\label{sec_optimization_analysis}
In this section,
we further discuss the implementation details of optimizing the proposed framework,
and analysis the optimization procedure with different settings.

\paragraph{Implementation Details}
As for the training on CIFAR-10 and CIFAR-100,
the learning rate for Eq. \ref{eq:interpret-KD-further-revised-solved} is set as 0.1,
and was changed to 0.01, and 0.001 at two steps (30k and 48k) respectively.
The optimizer for Eq. \ref{eq:interpret-KD-revised-split2}
started at a smaller learning rate 0.01,
but also is reduced according to similar strategies.
For the ILSVRC 2012,
the learning rate for Eq. \ref{eq:interpret-KD-further-revised-solved} is set as 0.1
with a ploy decreasing in each 6 epoch,
and the optimizer for Eq. \ref{eq:interpret-KD-revised-split2}
started at learning rate 0.005.
The weight decay of 0.00001 and momentum of 0.9 are all used.
For the works related to quantization strategies \cite{jacob2018quantization, rastegari2016xnor},
we try to evaluate the performance if combining these works with our framework.
Since the quantization techniques transfer the parameters distribution into a discrete space,
we found the optimization will be seriously impacted
and convergence performance also be influenced.
However,
this analysis is out of the scope of this paper,
so it is left as future work.

\paragraph{Joint Optimization}
For optimizing the $\tilde{W^t}$ with $W^t$ by Eq. \ref{eq:interpret-KD-further-revised-solved}
and the $W^s$ with $\tilde{W^t}$ by Eq. \ref{eq:interpret-KD-revised-split2},
we use two different optimizers to separately training these two sub-objective functions.
Moreover,
we tried different initialization techniques for parameters,
and we found the objective function is harder to converge,
if the initialization on $\tilde{W^t}$ is very different from $W^s$.
We also consider the types for different optimizers \cite{adam2014, Bottou2012, rmsprop2012}.
Through changing the two optimizers as Adam \cite{adam2014} or RMS \cite{rmsprop2012},
we found it caused a performance oscillation but less than $1\%$.

\section{Conclusion}
In this paper,
we propose a knowledge representing (KR) framework
mainly focusing on modeling the parameters distribution as prior knowledge.
We suggest a knowledge aggregation scheme to
represent the parameters knowledge from teacher network into more abstract level,
for alleviating the phenomenon of residual accumulation in the deeper layers.
We also design a sparse recoding penalty for constraining
the student network to learn with the penalized gradients.
It helps the student network to avoid
the over-regularization during knowledge distilling and converge faster.
In conclusion,
the proposed framework can prompt the student network
to preserve the key features of teacher network,
even though the student network does not have a strong representation ability.

\paragraph{Acknowledgements.} We thanks all reviewers for providing the constructive suggestions.

{\small
\bibliographystyle{ieee_fullname}
\bibliography{egbib}
}

\end{document}